\pdfoutput=1
\documentclass[article]{IEEEtran}
\usepackage{graphicx}
\usepackage[table,xcdraw]{xcolor}
\usepackage{adjustbox}
\usepackage{tabularx, booktabs}
\usepackage{tablefootnote}
\usepackage{array}
\usepackage{tabularx}

\usepackage{footnote}
\usepackage{multicol}
\usepackage{supertabular}
\usepackage{fancyhdr}
\usepackage{xcolor}
\usepackage{capt-of}
\usepackage{float}
\usepackage[linesnumbered,lined,ruled,commentsnumbered]{algorithm2e}
\usepackage{placeins}
\usepackage{afterpage}

\usepackage[space]{cite}

\usepackage{tikz}
\usepackage[cmex10]{amsmath}
\usepackage{amsfonts}
\usepackage[noend]{algpseudocode}
\usepackage{array}

\usepackage{mdwmath}

\usepackage{eqparbox}

\usepackage[tight,footnotesize]{subfigure}

\usepackage[]{caption}
\usepackage{hyperref}
\hypersetup{
    colorlinks=false,
    linkcolor=black,
    filecolor=magenta,      
    urlcolor=black,
    citecolor=blue,
}
\begin{document}

%
\title{Adversarial Attacks on Time Series}
%
%
\makeatletter
\let\@fnsymbol\@arabic
\makeatother

\author{{Fazle~Karim$^{1}$,~\IEEEmembership{Graduate Student Member,~IEEE}
        Somshubra~Majumdar$^{2}$,
        and~Houshang~Darabi$^{1}$,~\IEEEmembership{Senior Member,~IEEE}
        }
\thanks{$^{1}$Mechanical and Industrial Engineering, University of Illinois at Chicago, Chicago,IL}
\thanks{$^{2}$Computer Science, University of Illinois at Chicago, Chicago, IL}}
\maketitle

\begin{abstract}
Time series classification models have been garnering significant importance in the research community. However, not much research has been done on generating adversarial samples for these models. These adversarial samples can become a security concern. In this paper, we propose utilizing an adversarial transformation network (ATN) on a distilled model to attack various time series classification models. The proposed attack on the classification model utilizes a distilled model as a surrogate that mimics the behavior of the attacked classical time series classification models. Our proposed methodology is applied onto 1-Nearest Neighbor Dynamic Time Warping (1-NN ) DTW, a Fully Connected Network and a Fully Convolutional Network (FCN), all of which are trained on 42 University of California Riverside (UCR) datasets. In this paper, we show both models were susceptible to attacks on all 42 datasets. To the best of our knowledge, such an attack on time series classification models has never been done before. Finally, we recommend future researchers that develop time series classification models to incorporating adversarial data samples into their training data sets to improve resilience on adversarial samples and to consider model robustness as an evaluative metric. 
\end{abstract}


%
\IEEEpeerreviewmaketitle

\section{Introduction}

\bstctlcite{IEEEexample:BSTcontrol}

Over the past decade, machine learning and deep learning have been powering several aspects of society. \cite{lecun2015deep} Machine learning and deep learning are being used in some areas such as web searches \cite{boyan1996machine}, recommendation systems \cite{pazzani2007content}, and wearables \cite{ravi2017deep}. With the advent of smart sensors,  advancements in data collection and storage at vast scales, ease of data analytics and predictive modelling, time series data being collected from various sensors can be analyzed to determine regular patterns that are interpretable and exploitable. Classifying these time series data has been an area of interest by several researchers \cite{sirisambhand2019dimensionality, sharabiani2017efficient,sharabiani2018asymptotic,xi2006fast}. Time series classification models are being used in health care, where ECG data are used to detect patients with severe cognitive defects, in audio, where words are categorized into different phenomes, and in gesture recognition, where motion data is used to categorize action being made. Sensor data for resource and safety-critical applications such as manufacturing plants, industrial engineering, and chemical compound synthesis, when augmented by on-device analytics would allow automated response to avert significant issues in normal operation \cite{UCRArchive2018}.  A successful time series classification model is able to capture and generalize the pattern of time series signals such that it is able to classify unseen data. Similarly, computer vision classification models exploit the spatial structure intrinsic to images obtained in the real world. However, computer vision models have been shown to make incorrect predictions, on the seemingly correct input, which is termed as an adversarial attack.  Utilizing a variety of adversarial attacks, complex models are tricked to incorrectly predict the wrong class label. This is a serious security issue in neural networks widely used for vision-based tasks where adding slight perturbations or carefully crafted noise on an input image can mislead the image classification algorithm to make highly confident, yet wildly inaccurate predictions. \cite{mopuri2018generalizable, zhang2018anticipating} This has been a growing concern in the Computer Vision field, where Deep Neural Networks (DNN) have been shown to be particularly susceptible to attacks. \cite{oregi2018adversarial, song2018mat} While DNNs are state-of-the-art models for a variety of classification tasks in several fields, including time series classification \cite{fawaz2018deep,karim2018lstm,karim2018multivariate}, these vulnerabilities harmfully impact real-world applicability in domains where secure and dependable predictions are of paramount importance.\cite{miyato2018virtual,mopuri2018generalizable} Compounding the severity of this issue, recent work by Papernot et al has shown that adversarial attacks on a particular computer vision classifier can easily be transferred into other similar classifiers. \cite{papernot2016transferability} Only recently has the focus of attacks been shifted to Time Series classification models based on deep neural networks and classical models. \cite{oregi2018adversarial}

Several adversarial sample crafting techniques have been proposed to trick various image classification models that rely on DNN (state-of-the-art models for computer vision). Most of these techniques target the gradient information from the DNNs, which make them susceptible to these attacks. \cite{akhtar2018threat, madry2017towards,tramer2017ensemble} Generating adversarial samples for time series classification model has not been focused on, albeit the potentially large security risk they may possess. One major security concern exists in voice recognition tasks that convert text-to-speech. Carlini and Wagner \cite{carlini2018audio} show how text-to-speech classifiers can be attacked. In addition, they provide various audio clips where a text-to-speech classifier, DeepSpeech, is not able to correctly detect the speech. Other security concerns can exist in healthcare devices that use time series classification algorithms, where it can be tricked into misdiagnosing patients that can affect the diagnosis of their ailment. Time series classification algorithms used to detect and monitor seismic activity can be manipulated to create fear and hysteria in our society. Wearables that use time series data to classify activity of the wearer can be fooled into convincing the users they are doing other actions.  Most of the current state-of-the-art time series classification algorithms are classical approaches, such as 1 Nearest Neighbor - Dynamic Time Warping (1NN-DTW) \cite{keogh2005exact}, Kernel SVMs \cite{kampouraki2009heartbeat}, or sophisticated methods such as Weasel \cite{schafer2017fast}, COTE \cite{bagnall2015time}, Fast-Shapelet \cite{rakthanmanon2013fast}. However, DNNs are fast becoming excellent time series classifiers due to their simplicity and effectiveness. The traditional time series classification models are harder to attack as it can be considered a black-box model with a non-differentiable internal computation. As such, no gradient information can be exploited. However, DNN models are more susceptible to white-box attacks as their gradient information can easily be exploited. A white-box attack is where the adversary is ``given access to all elements of the training procedure'' \cite{tramer2017ensemble} - which includes the training dataset, training algorithm, the parameters and weights of the model, and the model architecture itself \cite{tramer2017ensemble}. Conversely, a black-box attack only has access to the target models training procedure and model architecture. \cite{tramer2017ensemble} In this paper, we propose a black-box and a white-box attack that can attack both classical and deep learning time series classification state-of-the-art models.  

In this work, we propose a proxy attack strategy on a target classifier via a student model, trained using standard model distillation techniques to mimic the behavior of the target classical time series classification models. The ``student'' network is the neural network distilled from another time series classification model, called the ``teacher'' model, that learns to approximate the output of the teacher model. Once the student model has been trained, our proposed adversarial transformation network (ATN) can then be trained to attack this student model. We apply our methodologies onto 1-NN DTW,  Fully Connected Network and Fully Convolutional Network (FCN) that are trained on 42 University of California Riverside (UCR) datasets.\cite{UCRArchive2018} To the best of our knowledge, the result of such an attack on time series classification models has never been studied before. Finally, we recommend future researchers that develop time series classification models to consider model robustness as an evaluative metric and incorporate adversarial data samples into their training data sets in order to further improve resilience to adversarial attacks. 


The remainder of this paper is structured as follows: Section \ref{Background Works} provides a brief background on a couple time series classification models and background information on a few adversarial crafting techniques used on computer vision problems. Section \ref{Methodology} details our proposed methodologies and Section \ref{Experiments} presents and explains the results of our proposed methodologies on a couple of time series classification models. Section \ref{conclusion} concludes the paper and proposes future work.

\section{Background \& Related Works}
\label{Background Works}

\subsection{Time Series Classification Models}
\label{TSClassifiers}

\subsubsection{1-NN Dynamic Time Warping} 

\par The equations and definitions below are obtained from Kate et al. \cite{kate2016using} and Xi et al.  \cite{xi2006fast}. Dynamic Time Warping is a measures of similarity between 2 time series, $Q$ and $C$, which is detected by finding their best alignment. Time series $Q$ and $C$ are defined as:
\begin{align}
\label{eq:1}
Q &= q_1, q_2, q_3, ..., q_i, ..., q_n \\
\label{eq:2}
C &= c_1, c_2, c_3, ..., c_i, ..., c_n.
\end{align}

To align both the time series data, the distance between each timestep of $Q$ and $C$ is calculated, $(q_i - c_j)^2$, to generate a $n$-by-$m$ matrix. In other words, the $i^{\text{th}}$ and $j^{\text{th}}$ of the matrix is the $q_i$ and $c_j$. The optimal alignment between $Q$ and $C$ is considered the warping path, $W$, such that $W = w_1, w_2, w_3, ..., w_k, ..., w_K$. The warping path is computed such that,
\begin{enumerate}
\item $w_1 = (1,1)$,
\item $w_k=(n,m)_k$,
\item given $w_k = (a, b)$ then $w_{k - 1} = (a^{\prime}, b^{\prime})$ where $0 \leq a - a^{\prime} \leq 1$ and $0 \leq b - b^{\prime} \leq 1$.
\end{enumerate}

The optimal alignment is the warping path that minimizes the total distance between the aligning points,

\begin{equation}
\label{eq:3}
DTW(Q,C) = \operatorname*{argmin}_{W=w_1,w_2,...,w_K}\sqrt{\sum_{k=1,w_k=(i,j)}^{k} (q_i - c_j)^2}.
\end{equation}

\subsubsection{Fully Convolutional Network} 

The Fully Convolutional Network (FCN) is one of the earliest deep learning time series classifier. \cite{wang2017time} It contains 3 convolutional layers, with convolution kernels of size 8, 5 and 3 respectively, and emitting 128, 256 and 128 filters respectively. Each convolution layer is followed by a batch normalization \cite{ioffe2015batch} layer that is applied with a ReLU activation layer. A global average pooling layer is employed after the last ReLU activation layer. Finally, softmax is applied to determine the class probability vector. 

\subsection{Adversarial Transformation Network}
\label{AdversarialTransformNetwork}

Several methods have been proposed to generate adversarial samples that attack deep neural networks that are trained for computer vision tasks. Most of these methods use the gradient with respect to the image pixels of these neural networks. Baluja and Fischer \cite{baluja2017adversarial} propose Adversarial Transformation Networks (ATNs) to efficiently generate an adversarial sample that attacks various networks by training a feed-forward neural network in a self-supervised method. Given the original input sample, ATNs modify the classifier outputs slightly to match the adversarial target. ATN works similarly to the generator model in the Generative Adversarial Training framework. 

According to Baluja and Fischer et al. \cite{baluja2017adversarial}, an ATN can be parametrize as a neural network ${g_f (x) : x \rightarrow \hat{x}}$, where $f$ is the target model (either a classical model or another neural network) which outputs either a probability distribution across class labels or a sparse class label, and $\hat{x} \sim x$, but argmax $f(x)$ $\neq$ argmax $f(\hat{x})$. To find $g_f$, we minimize the following loss function :
\begin{equation}
    L = \beta * L_x (g_f (\textbf{x}_i), \textbf{x}_i) + L_y (f(g_f (\textbf{x}_i)), f(\textbf{x}_i))
\end{equation}
where $L_x$ is a loss function on the input space (e.g. $L_2$ loss function), $L_y$ is the specially constructed loss function on the output space of $f$ to avoid learning the identity function, $\textbf{x}_i$ is the i-th sample in the dataset and $\beta$ is the weighing term between the two loss functions. 

It is necessary to carefully select the loss function $L_y$ on the output space to successfully avoid learning the identity function. Baluja and Fischer et al. \cite{baluja2017adversarial} define the loss function $L_y$ as $L_y (\textbf{y}', \textbf{y}) = L_2 (\textbf{y}', r(\textbf{y}, t))$, where $\textbf{y} = f(x)$, $\textbf{y}' = f(g_f (x))$ and $r(\cdot)$ is a reranking function that modifies \textbf{y} such that $y_k < y_t, \forall k \neq t$. This reranking function $r(\textbf{y}, t)$ can either be the simple one hot encoding function $onehot(t)$ or be formulated to take advantage of the already present $\textbf{y}$ to encourage better reconstruction. We therefore utilize the reranking function proposed by Baluja and Fischer et al. \cite{baluja2017adversarial}, which can be formulated as:
\begin{equation}
    r_\alpha (\textbf{y}, t) = norm  \left( \left.
    \begin{cases}  
    \alpha * max \: y,& \text{if } k = t\\
    y_k,& \text{otherwise}
    \end{cases}
    \right \}_{k \in y}
    \right)
\end{equation}
where $\alpha > 1$ is an additional hyper parameter which defines how much larger $y_t$ should be than the current max classification and $norm$ is a normalizing function that rescales its input to be a valid probability distribution

\subsection{Transferability Property}

Papernot et al. \cite{papernot2016transferability} propose a black-box attack by training a local substitute network, $s$, to replicate or approximate the target DNN model, $f$. The local substitute model is trained using synthetically generated samples and the output of these samples are labels from $f$. Subsequently, $s$ is used to generate adversarial samples that it misclassifies. Generating adversarial samples for $s$ is much easier, as its full knowledge/parameters are available, making it susceptible to various attacks. The key criteria to successfully generate adversarial samples of $f$ is the transferability property, where adversarial samples that misclassify $s$ will also misclassify $f$. 

\subsection{Knowledge Distilation}

Knowledge distillation, first proposed by Bucila et al. \cite{bucilua2006model}, is a model compression technique where a small model, $s$, is trained to mimic a pre-trained model, $f$. This process is also known as the model distillation training where the teacher is $f$ and the student is $s$. The knowledge that is distilled from the teacher model to the student model is done by minimizing a loss function, where the objective of the student model is to imitate the distribution of the class probabilities of the teacher model. Hinton et al. \cite{hinton2015distilling} note that there are several instances where the probability distribution is skewed such that the correct class probability would have a probability close to 1 and the remaining classes would have a probability closer to 0. Hence, Hinton et al. \cite{hinton2015distilling} recommend computing the probabilities $q_i$ from the pre-normalized logits $z_i$, such that:
\begin{equation}\label{eq:scaled-softmax}
    q_i = \sigma(z; T) = \frac{exp \: (z_i / T)}{\sum_j exp \: (z_j / T)}
\end{equation}
where $T$ is a temperature factor normally set to 1. Higher values of $T$ produce softer probability distributions over classes. The loss that is minimized is the model distillation loss, further explained in Section \ref{sec:TrainingMethodology}.

\section{Methodology}
\label{Methodology}

\subsection{Gradient Adversarial Transformation Network}
In this work, we employ distinct methodologies for white-box and black-box attacks, in order to adhere to a strictly realistic set of limitations in black-box attacks. For both methodologies, we incorporate Adversarial Transformation Networks (ATN) \cite{baluja2017adversarial} as a generative neural network that accepts an input time series sample $x$ and transforms it to an adversarial sample $\hat{x}$. 

An Adversarial Transformation Network can be formulated as a neural network $g_f (x): x \rightarrow \hat{x}$, where $f$ is the model that will be attacked. We further augment the information available to the ATN with the gradient of the input sample $x$ with respect to the softmax scaled logits of the target class predicted by the attacked classifier. We can therefore formally define a Gradient Adversarial Transformation Network (GATN) as a neural network parametrized as $g_f (x, \tilde{x}): (x, \tilde{x}) \rightarrow \hat{x}$, where :
\begin{equation}
    \tilde{x} = \frac{\partial \, x}{\partial \, f_{t}}
\end{equation}
such that $x \in \mathbb{R}^T$ is an input time series of maximum length $T$, $f_{t}$ represents the probability of the input time series being classified as the target class, $t$. With the availability of the input gradient $\tilde{x}$, the Gradient Adversarial Transformation Network can better construct adversarial samples that can affect the targeted model while reducing the overall perturbation added to the sample. Therefore we utilize the GATN model for all of our attacks. 

A significant issue with the above formulation of the GATN is the non-differentiability of classical models. Distance-based models such as 1-NN Dynamic Time Warping does not have the notion of gradients during either training or evaluation, and therefore we cannot directly compute the gradient of the input ($\tilde{x}$) with respect to the 1-NN DTW model $f$. We discuss the solution to this issue in Section \ref{sec:TrainingMethodology}, by building a student neural network $s$ which approximates the predictions of the non-differentiable classical classifier $f$.

\begin{figure*}[htpb]
\centering
\fbox{
\includegraphics[width=0.5\linewidth]{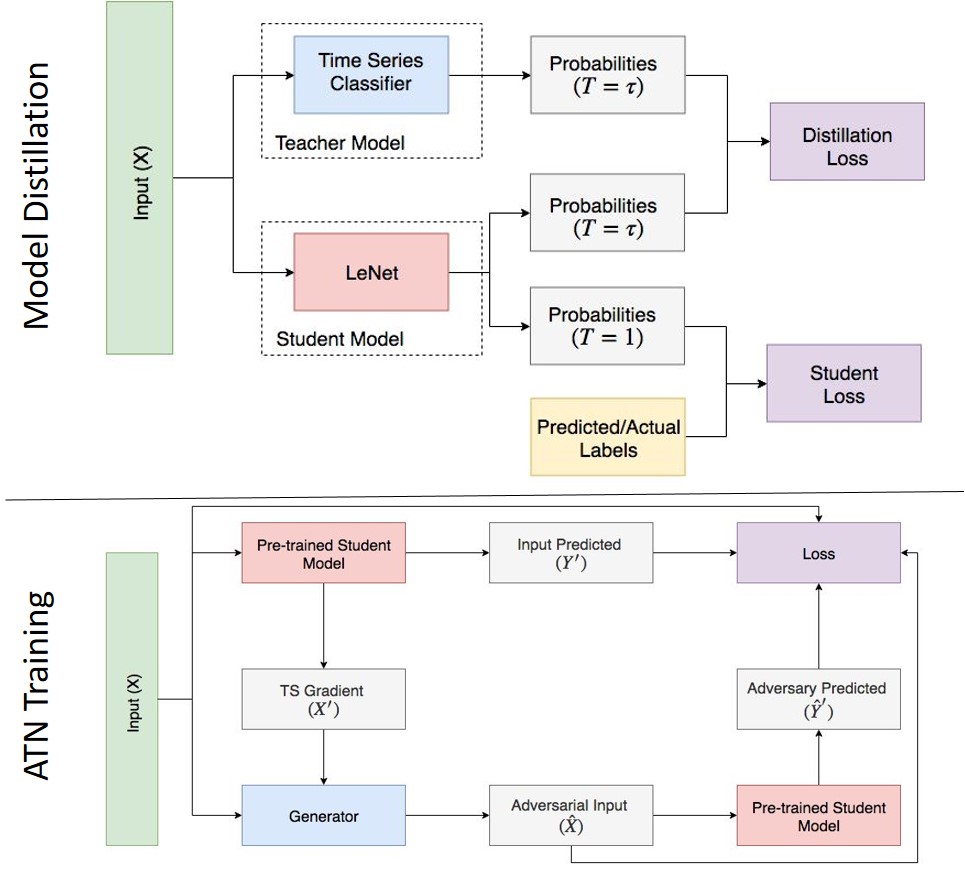}
}
\caption{The top diagram shows the methodology of training the model distillation used in the white-box and black-box attacks. The bottom diagram is the methodology utilized to attack a time series classifier. }
\label{fig_distilation1}

\end{figure*}

\subsection{Black-box \& White-box Restrictions}
\label{BlackWhiteBoxRestrictions}

While this formulation of the GATN is sufficient for white-box attacks where we have access to the attacked model $f$ or the student model $s$, this assumption is unrealistic in the case of black-box attacks. For a black-box, we are not permitted access to either the internal model (a neural network or a classical model) or to the dataset that the model was trained on. Furthermore, for black-box attacks, we impose a restriction on the predicted labels, such that we utilize only the class label predicted, and not the probability distribution produced after softmax scaling (for neural networks), or the scaled probabilistic approximations of classical model predictions. 

To further restrict ourselves to realistic attack vectors, we stratify the available dataset $D$, which will be used to train the GATN, into two halves, such that we train the GATN on one subset, $D_{eval}$, and are able to perform evaluations on both this train set and the wholly unseen test set, $D_{test}$. Note that this available dataset $D$ is not the dataset on which the attacked model $f$ was trained on. As such, we never utilize the train set available to the attacked classifier to either train or evaluate the GATN model. In order to satisfy these constraints on available data, we define our available dataset $D$ as the test set of the UCR Archive \cite{UCRArchive2018}. As the test set was never used to train any atacked model $f$, it is sufficient to utilize it as an unseen dataset. We then split the test dataset into two class-balanced halves, $D_{eval}$ and $D_{test}$. Another convenience is the availability of test set labels, which can be harnessed as a strict check when evaluating adversarial generators.

When we evaluate under the constraints of black-boxes, we further limit ourselves to “unlabeled” train sets, where we assume the available dataset is unlabeled, and thereby utilize only the predicted label from the attacked classifier $f$ to label the dataset prior to attacks. We state this as an important restriction, considering that it is far more difficult to freely obtain or create datasets for time series than for images which are easily understood and interpreted. For time series, significant expertise may be required to distinguish one sample amongst multiple classes, whereas natural images can be coarsely labeled with relative ease without sophisticated equipment or expertise.


\subsection{Training Methodology}
\label{sec:TrainingMethodology}
A chief consideration during training of ATN or GATN is the loss formulation on the prediction space ($L_y$) is heavily influenced by the reranking function $r(\cdot)$ chosen. If we opt for the one hot encoding of the target class, we lose the ability to maintain class ordering and the ability to adjust the ranking weight ($\alpha$) to obtain adversaries with less distortion. However, to utilize the appropriate reranking function, we must have access to the class probability distribution, which is unavailable to black-box attacks, or may not even be possible to compute for certain classical models such as 1-NN DTW which uses distance-based computations to determine the nearest neighbor. 

To overcome this limitation, we employ knowledge distillation as a mechanism to train a student neural network $s$, which is trained to replicate the predictions of the attacked model $f$. As such, we are required to compute the predictions of the attacked model on the dataset we possess just one time, which can be either class labels or probability distribution over all classes. We then utilize these labels as the ground truth labels that the student $s$ is trained to imitate. In case the predictions are class labels, we utilize one hot encoding scheme to compute the cross entropy loss, otherwise, we try to imitate the probability distribution directly. It is to be noted that the student model shares the training dataset $D_{eval}$ with the GATN model. 

As suggested by Hinton et al. \cite{hinton2015distilling}, we describe the training scheme of the student as shown in Figure \ref{fig_distilation1}. We scale the logits of the student $s$ and teacher $f$ (iff the teacher provides probabilities and it is a white-box attack) by a temperature scaling parameter $\tau$, which is kept constant at 10 for all experiments. When training the student model, we minimize the loss function defined as:
\begin{align}
    L_{transfer} &= \gamma * L_{distillation} + (1 - \gamma) * L_{student}\\
L_{distillation} &= \mathcal{H}(\sigma(z_f; T=\tau), \, \sigma(z_s; T=\tau))\\
    L_{student} &= \mathcal{H}(y,\sigma(z_s; T=1))
\end{align}
where $\mathcal{H}$ is the standard cross entropy loss function, $z_s$ and $z_f$ are the un-normalized logits of the student ($s$) and teacher ($f$) models respectively, $\sigma(\cdot)$ is the scaled-softmax operation described in Equation (\ref{eq:scaled-softmax}), $y$ is the ground truth labels, and $\gamma$ is a gating parameter between the two losses and is used to maintain a balance between how much the student $s$ imitates the teacher $f$ versus how much it learns from the hard label loss. When training a student as a white-box attack, we set $\gamma$ to be 0.5, allowing the equal weight to both losses, whereas for a black-box attack, we set $\gamma$ to be 1. Therefore for black-box attacks, we force the student $s$ to only mimic the teacher $f$ to the limit of its capacity. In setting this restriction, we limit the amount of information that may be made available to the GATN.

Once we have a student model $s$ which is capable of simulating the predictions of the attacked model $f$, we then train the GATN using this student model. Figure \ref{fig_distilation1} shows the methodology of training such a model. Since the GATN requires not just the original sample $x$ but also the gradient of that sample $\tilde{x}$ with respect to the predictions for the targetted class, we require two forward passes from the student model. The first forward pass is simply to obtain the gradient of the input $\tilde{x}$, as well as the predicted probability distribution of the student $y’$. The adversarial sample crafted ($\hat{X}$) is then used in a second forward pass to compute the predicted probability distribution of the student with respect to the adversarial sample, $\hat{y}'$.  We minimize the weighted loss measure $L$ defined in Section \ref{AdversarialTransformNetwork} in order to train the GATN model. 

\subsection{Evaluation Methodology}

Due to the different restrictions imposed between available information depending on whether the attack is a white-box or black-box attack, we train the GATN on one of two models. We assert that we train the GATN by attacking the target neural network $f$ directly only when we perform a white-box attack on a neural network. In all other cases, whether the attack is a white-box or black-box attack, and whether the attacked model is a neural network or a classical model, we select the student model $s$ as the model which is attacked to train the GATN, and then use the GATN's predictions ($\hat{x}$) to check if the teacher model $f$ is also attacked when provided the predicted adversarial input ($\hat{x}$) as a sample.

During evaluation of the trained GATN, we compute the number of adversaries of the attacked model $f$ that have been obtained on the training set $D_{eval}$. During the evaluation, we can measure any metric under two circumstances. Provided a labeled dataset which was split, we can perform a two-fold verification of whether an adversary was found or not. First, we check that the ground truth label matches the predicted label of the classifier when provided with an unmodified input ($y = y'$ when input $x$ if provided to $f$), and then check whether this predicted label is different from the predicted label when provided with the adversarial input ($y’ \neq \hat{y}'$ when input $\hat{x}$ is provided to $f$). This ensures that we do not count an incorrect prediction from a random classifier as an attack. 

Another circumstance is that we do not have any labeled samples prior to splitting the dataset. This training set is an unseen set for the attacked model $f$, therefore we consider that the dataset is “unlabeled”, and assume that the label predicted by the base classifier is the ground truth ($y = y' $ by default, when sample $x$ is provided to $f$). This is done prior to any attack by the GATN and is computed just once. We then define an adversarial sample as a sample $\hat{x}$ whose predicted class label is different than the predicted ground truth label ($y’ \neq \hat{y}'$, when sample $\hat{x}$ is provided to $f$). A drawback of this approach is that it is overly optimistic and rewards sensitive classifiers that misclassify due to very minor alterations.

In order to adhere to an unbiased evaluation, we chose the first option, and utilize the provided labels that we know from the test set to properly evaluate the adversarial inputs. In doing so, we acknowledge the necessity of a labeled test set, but as shown above, it is not strictly necessary to follow this approach.


\section{Experiments}
\label{Experiments}
All methodologies were tested on 42 benchmark datasets for time series classification found in the UCR repository. The 42 datasets selected were all from the types ``Sensor'', ``ECG'', ``EOG'', and ``Hemodynamics'', where an adversarial attack is a potential security concern. We evaluate based on two criterion, the mean squared error between the training dataset and the generated samples (lower is better) and; the number of adversaries for a set of chosen beta values (higher is better). For all experiments, we keep $\alpha$, the reranking weight, set to 1.5, the target class set to 1, and perform a grid search over 5 possible values of $\beta$, the reconstruction weight term, such that $\beta = 10^{-b}; \:\: b \in \{1,2,3,4,5\}$. The codes and weights of all models are available at \href{https://github.com/houshd/TS_Adv}{https://github.com/houshd/TS\_Adv}

\subsection{Experiments}
We select both neural networks as well as traditional models as the attacked model $f$. For the attacked neural network, we utilize a Fully Convolutional Network, whereas for the base traditional model, 1NN-Dynamic Time Warping Classifier is utilized. 

To maintain the strictest definition of the black and white-box attacks, we utilize only the discrete class label of the attacked model for black-box attacks and utilize the probability distribution predicted by the classifier for white-box attacks. The only exception where a student-teacher network is not used is when performing a white-box attack on a FCN time series model, as the gradient information of a neural network can be directly exploited by an Adversarial Transformation Network (ATN). The performance of the adversarial model is evaluated on the original time series classification “teacher” model.  

For every student model we train, we utilize the LeNet-5 architecture \cite{lecun2015lenet}. We define a LeNet-5 time series classifier as a classical Convolutional Neural network following the structure : Conv (6 filters, 5x5, valid padding) - Max Pooling - Conv (16 filters, 5x5, valid padding) - Max Pooling - Fully Connected (120 units, relu) - Fully Connected (84 units, relu) - Fully Connected (number of classes, softmax).

The fully convolutional network is based on the FCN model proposed by \textit{Wang et al} \cite{wang2017time}. It is comprised of 3 blocks, each comprised of a sequence of Convolution layer - Batch Normalization - ReLU activations. All convolutional kernels are initialized using the uniform he initialization proposed by He et al. \cite{he2015delving} We utilize [128, 256, 128] filters and kernel sizes of [8, 5, 3] to be consistent.

A strong determinisitic baseline model to classifiy time series is 1-NN DTW with 100\% warping window. Due to its reliance on a distance matrix as a means of its classification, it cannot easily be used to compute an equivalent soft probabilistic representation. Since white-box attacks have access to the probability distribution predicted for each sample, we utilize this distance matrix in the computation of an equivalent soft probabilistic representation. The equivalent representation is such that if we compute the top class (class with highest probability score) on this representation, we get the exact same result as selecting the 1-nearest neighbor on the actual distance matrix.

To compute this soft probabilistic representation, consider a distance matrix $V$ computed using a distance measure such as DTW between all possible pairs of samples between the two datasets being compared. 
\begin{algorithm}
  \TitleOfAlgo{Soft-1NN (V, y)}
  \DontPrintSemicolon
  \SetAlgoLined
  \KwData{V is a distance matrix of shape $[N_{test},N_{train}]$ and y is the train set label vector of length $N_{train}$}
  \KwResult{Softmax normalized predictions p of shape $[N_{test}, C]$ and the discrete label vector q of length $N_{test}$}
  
  \Begin{
    $V \longleftarrow {(-V)}$\;
    \BlankLine
    
    $Unique\_classes = Unique(y)$   // unique class labels\;
    $V_c$ = []\;
    
    \BlankLine
    \For{$c_i$ in Unique\_classes}{
        $v_c = V_{(y = c_i)}$  // $[N_{test}, N_{train}{(y = c_i)}]$\;  
        $v_c\_max = max(v_c)$  // $[N_{test}]$\; 
        $V_c$.append($v_c\_max$)\;
    }
    \BlankLine
    V' = concatenate($V_c$) // $[N_{test}, \text{number of classes]}$\;
    p = softmax(V') // $[N_{test}, \text{number of classes}]$\;
    q = argmax(p) // $[N_{test}]$ \;
    
    \BlankLine
    
    \Return (p, q)\;
  }
  \caption{Equivalent probabilistic representation of the distance matrix for 1-nearest neighbor classification}
  \label{algo1}
\end{algorithm}

Algorithm \ref{algo1} is an intermediate normalization algorithm which accepts a distance matrix $V$ and the class labels of the training set $y$ as inputs and computes an equivalent probabilistic representation that can directly be utilized to compute the 1-nearest neighbor. The $Soft{-1NN}$ algorithm selects all samples that belong to a class $c_i$, where $i \in \{1, \dots, C\}$  as $v_c$, computes the maximum over all train samples for that class and appends the vector $v_c\_max$ to the list $V_c$. The concatenation of all of these lists of vectors in $V_c$ then represents the matrix  $V'$, on which we then apply the $softmax$ function, as shown in Equation \ref{eq:scaled-softmax} with $T$ set to 1, to represent this matrix $V'$ as a probabilistic equivalent of the original distance matrix $V$.

An implicit restriction placed on Algorithm \ref{algo1} is that the representation is equivalent only when computing the 1-nearest neighbor. It cannot be used to to represent the $K$-nearest neighbors and therefore cannot be used for $K$-nearest neighbor classification. However, in time series classification, the general consensus is on the use of 1-nearest neighbor classifiers and its variants to classify time series. \cite{keogh2005exact,sharabiani2018asymptotic,sharabiani2017efficient,bagnall2015time,schafer2015boss} While the above algorithm has currently been applied to convert the 1NN-DTW distance matrix, it can also be applied to normalize any distance matrix utilized for 1-nearest neighbors classification algorithms.



	


\subsection{Results}
\renewcommand{\floatpagefraction}{0.75}

\begin{figure*}[htpb]
\centering
\fbox{
\includegraphics[width=0.8\linewidth]{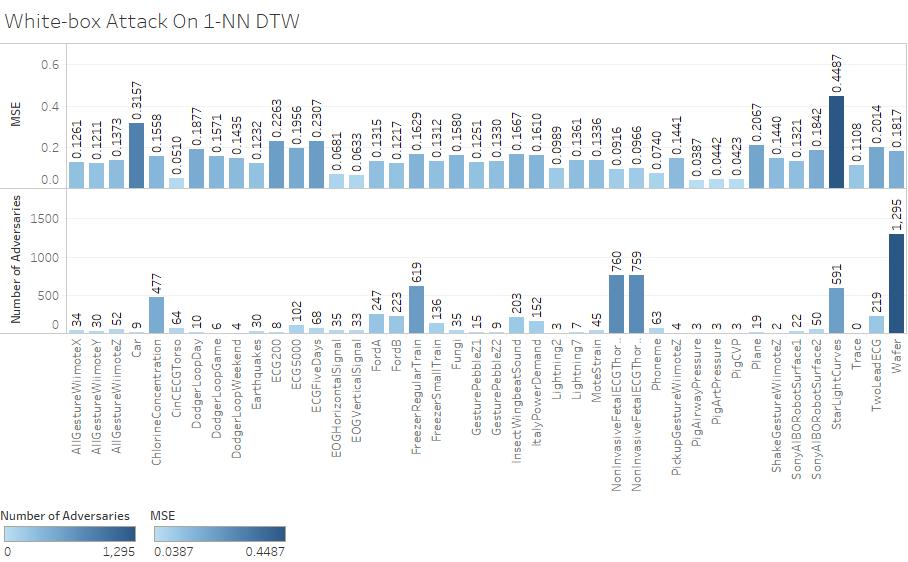}
}
\caption{White-box attack on 1-NN DTW that is trained on all 42 datasets}
\label{fig:wb_1nn}

\end{figure*}

\begin{figure*}[htpb]
\centering
\fbox{
\includegraphics[width=0.8\linewidth]{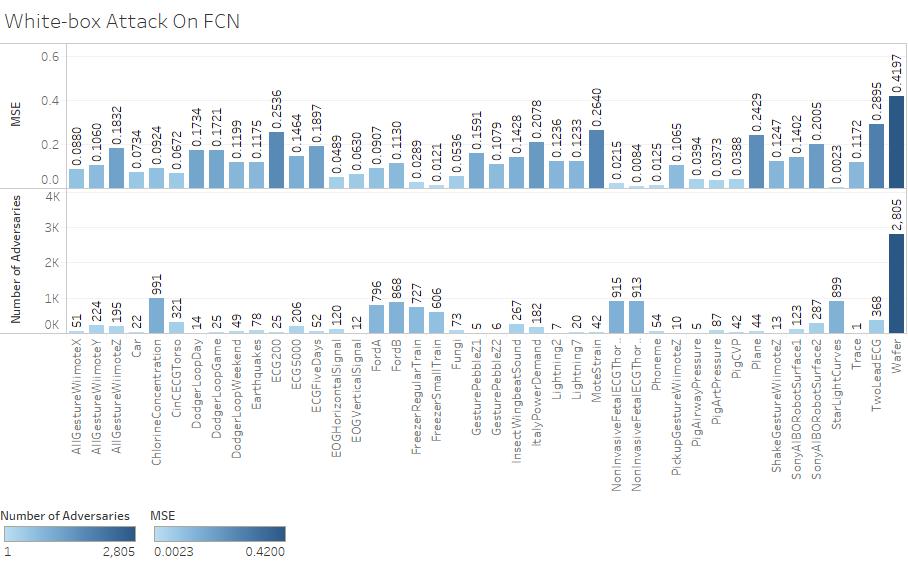}
}
\caption{White-box attack on FCN that is trained on all 42 datasets}
\label{fig:wb_fcn}

\end{figure*}

\begin{figure*}[htpb]
\centering
\fbox{
\includegraphics[width=0.8\linewidth]{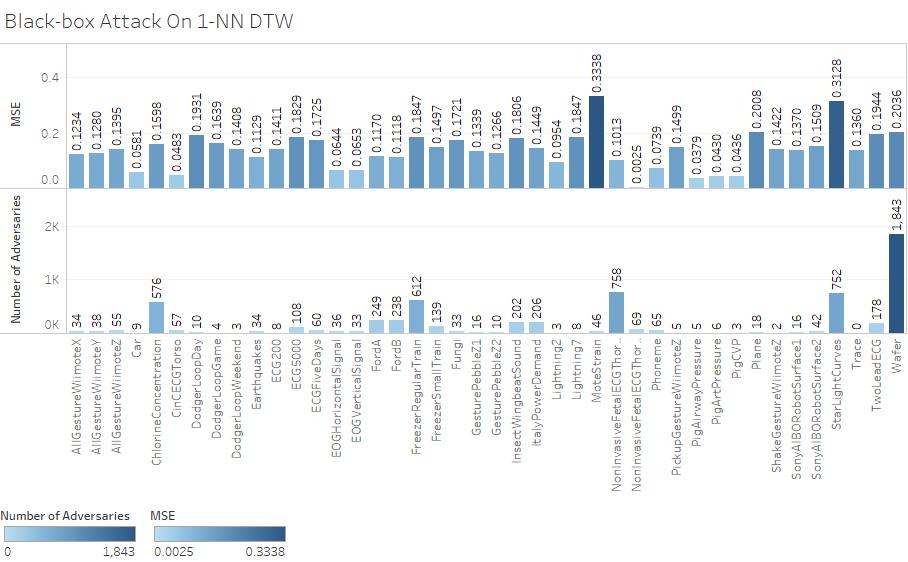}
}
\caption{Black-box attack on 1-NN DTW that is trained on all 42 datasets}
\label{fig:bb_1nn}

\end{figure*}

\begin{figure*}[htpb]
\centering
\fbox{
\includegraphics[width=0.8\linewidth]{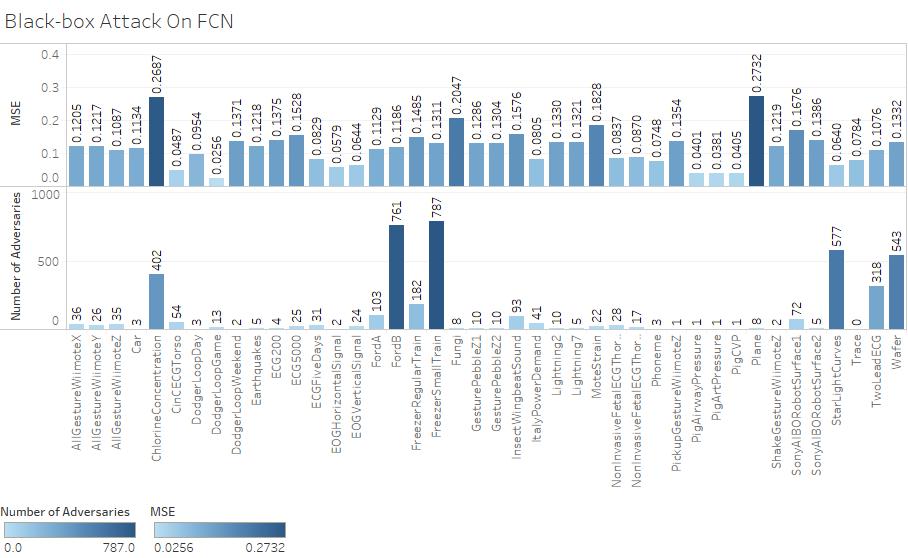}
}
\caption{Black-box attack on FCN that is trained on all 42 datasets}
\label{fig:bb_fcn}

\end{figure*}

\begin{figure*}[htpb]
\centering

\includegraphics[width=0.8\linewidth]{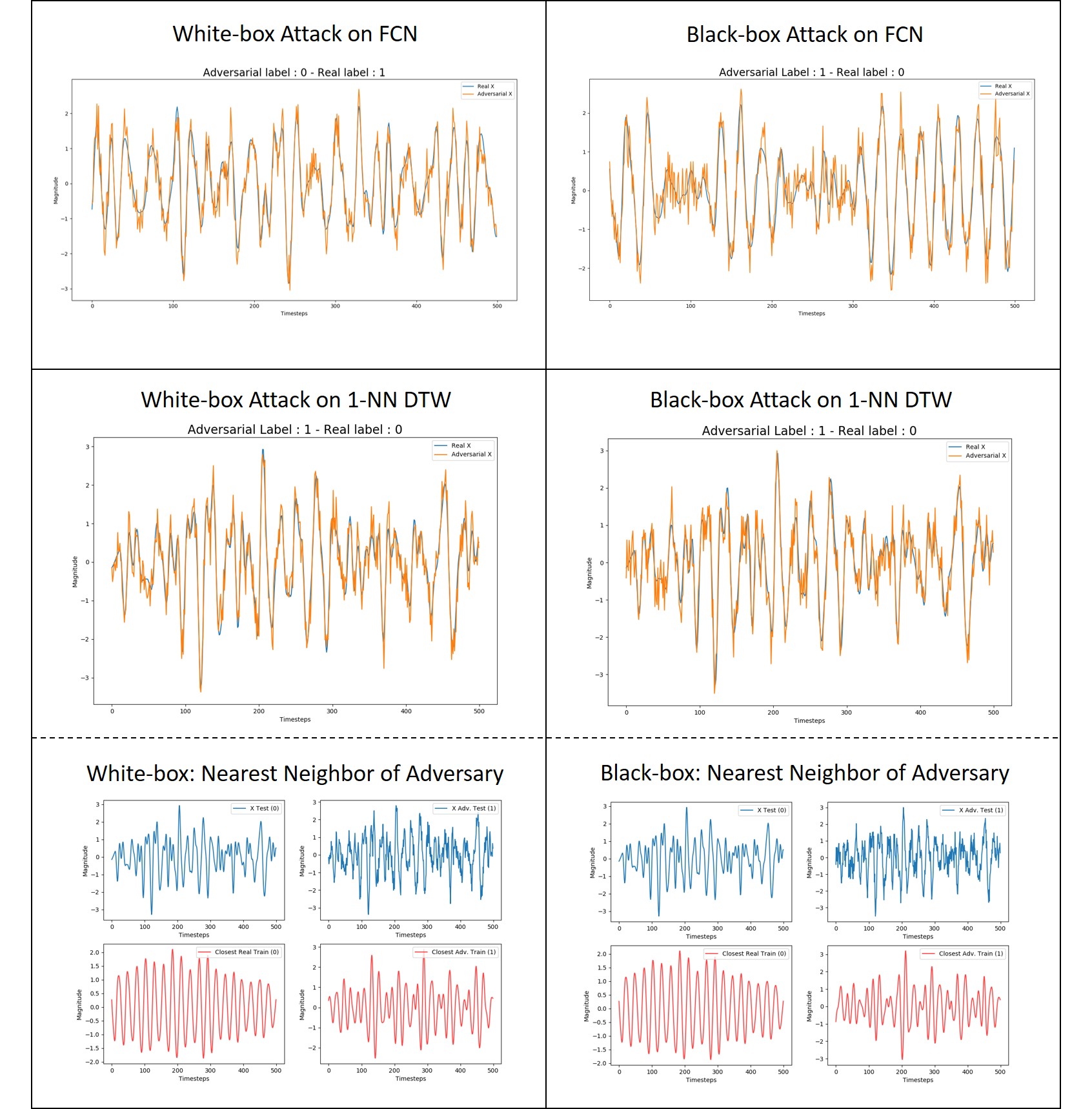}

\caption{A sample black-box and white-box attack on an FCN and 1-NN DTW classifier that is trained on the dataset ``ForbB''. The last row of the figure depicts the nearest neighbor of the original and adversarial time series.}
\label{fig:samp_adv}

\end{figure*}

Figures \ref{fig:wb_1nn} and \ref{fig:wb_fcn} depict the results from white-box attacks on 1-NN DTW and FCN that is applied on 42 UCR datasets. Further, Figures \ref{fig:bb_1nn} and \ref{fig:bb_fcn} represent the results from black-box attacks on 1-NN DTW and FCN classifiers that are trained on the same 42 UCR datasets. The detailed results can be found in Appendix A. The proposed methodology is successfully in capturing adversaries on all datasets. An example of an adversarial attack on the dataset ``FordB'' is shown in Figure \ref{fig:samp_adv}.  

The number of adversaries and amount of perturbation per sample in each dataset can increase or decrease depending on the hyper-parameters that are tested on. For example, the dataset ``Trace'' has 0 adversaries for most of the attacks (black-box attack on 1-NN DTW, white-box attack on 1-NN DTW, black-box attack on FCN) when the Target Class is set to 1. However, if the target class is changed to 2, the number of adversaries generated increases to 9,3,1,37 for a black-box attack on 1-NN DTW, white-box attack on 1-NN DTW, black-box attack on FCN and white-box attack on FCN, respectively. These numbers can be higher if the hyper-parameters are changed. In addition, due to the loss function of the ATN, the target class has a significant impact on the adversary being generated. It is easier to generate adversaries for time series classes that are similar to each other. 

\begin{table}[htbp]
  \centering
  \caption{Wilcoxson signed-rank test comparing the number of adversaries between the different attacks}
  \begin{adjustbox}{width=1 \linewidth}
    \begin{tabular}{|c|c|c|c|}
    \hline
          & White-box 1-NN DTW & Black-box FCN & White-box FCN \\
    \hline
    Black-box 1-NN DTW & 1.864E-01 & \cellcolor[rgb]{ .776,  .937,  .808}\textcolor[rgb]{ 0,  .38,  0}{6.706E-04} & \cellcolor[rgb]{ .776,  .937,  .808}\textcolor[rgb]{ 0,  .38,  0}{2.013E-06} \\
    \hline
    White-box 1-NN DTW & \cellcolor[rgb]{ 0,  0,  0} & \cellcolor[rgb]{ .776,  .937,  .808}\textcolor[rgb]{ 0,  .38,  0}{6.994E-04} & \cellcolor[rgb]{ .776,  .937,  .808}\textcolor[rgb]{ 0,  .38,  0}{6.681E-07} \\
    \hline
    Black-box FCN & \cellcolor[rgb]{ 0,  0,  0} & \cellcolor[rgb]{ 0,  0,  0} & \cellcolor[rgb]{ .776,  .937,  .808}\textcolor[rgb]{ 0,  .38,  0}{5.680E-07} \\
    \hline
    \end{tabular}%
    \end{adjustbox}
  \label{tab:num_pvalue}%
\end{table}%

A Wilcoxson signed-rank test is utilized to compare the number of adversaries generated by white-box and black-box attacks on FCN and 1-NN classifiers that are trained on the 42 datasets, summarized in Table \ref{tab:num_pvalue}. Our results indicate that the FCN  classifier is more susceptible to a white-box attack compared to a white-box attack on 1-NN DTW. It is to be noted that the white-box attack on the FCN classifier generates significantly more adversaries than its counterparts. This is because the white-box attack is directly on the FCN model and not on a student model that approximates the classifier behavior. We observe that the number of adversarial samples obtained from black-box attacks on FCN classifiers are greater than the number of adversarial samples from either white-box or black-box attacks on DTW classifiers. A Wilcoxson signed-rank test confirms this observation by showing a statistically significant difference in number of adversarial samples detected due to the black-box or white-box attacks on 1-NN DTW classifiers versus the number of adversarial samples obtained via black-box attacks on the FCN classifiers. In summary, we observe the largest number of adversarial samples for the FCN model when under a white-box attack. We also detect that 1-NN DTW classifiers under either attack have approximately the same number of adversarial samples. Finally, we discover that FCN has the least number of adversarial samples after black box attacks, though each of those samples requires indiscernible perturbations to the original signal. These observations are important for future researchers who develop time series classifiers, as the number of adversarial samples generated under each methodology can be used as an evaluative metric to measure the robustness of a model. 

The average MSE of adversarial samples after black-box attacks on FCN classifiers is significantly lower than the average MSE of the adversarial samples obtained via black-box and white-box attacks on 1-NN DTW classifiers, as observed in Table \ref{tab:mse_pvalue}. A lower MSE indicates the black-box attack on FCN classifiers requires minimal perturbations per time series sample in comparison to the attacks on 1-NN DTW classifiers.

\begin{figure*}[htpb]
\centering
\fbox{
\includegraphics[width=0.8\linewidth]{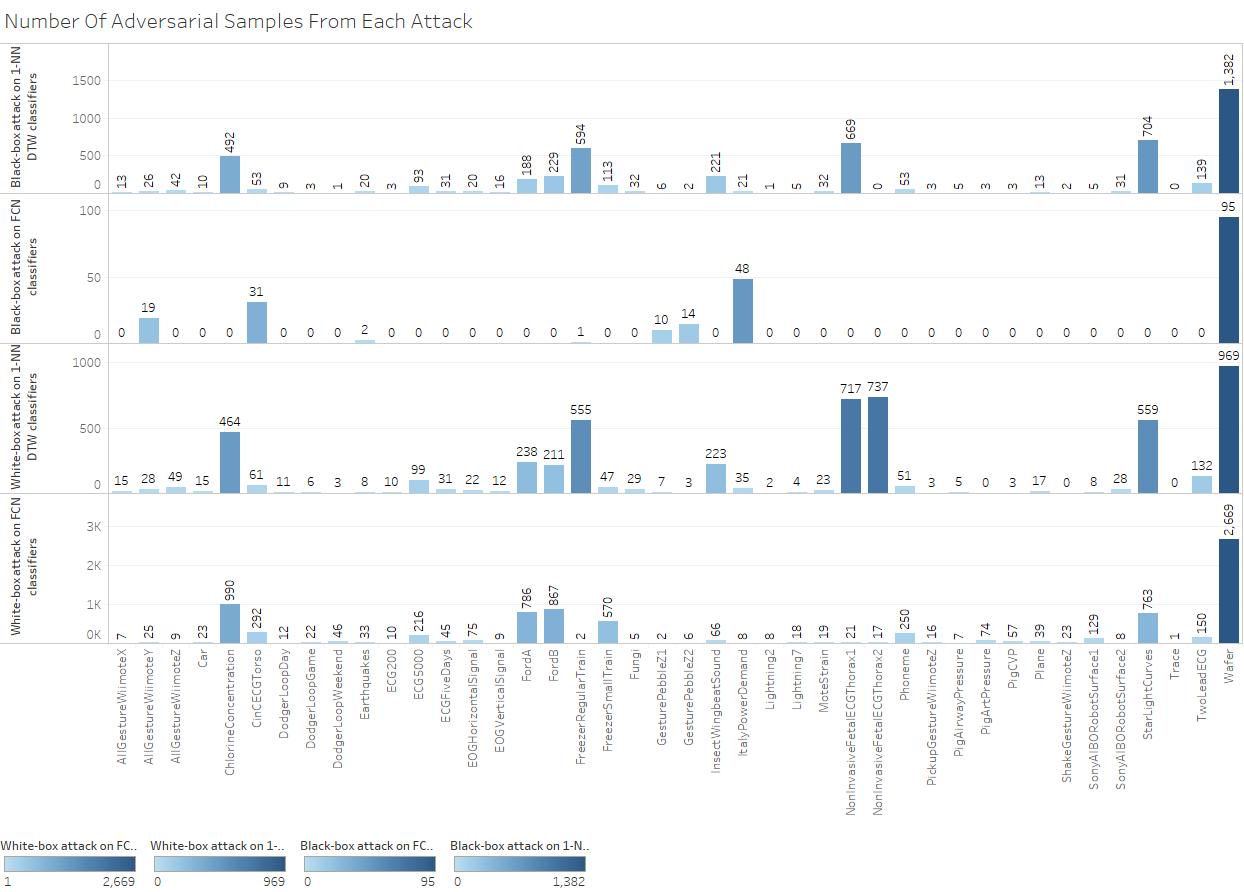}
}
\caption{Black-box and white-box attacks on FCN and 1-NN DTW classifiers that are tested on $D_{test}$ without any retraining.}
\label{fig:gen}

\end{figure*}

\begin{table}[htbp]
  \centering
  \caption{Wilcoxson signed-rank test comparing the MSE between the different attacks}
    \begin{adjustbox}{width=1 \linewidth}

    \begin{tabular}{|c|c|c|c|}
    \hline
          & White-box 1-NN DTW & Black-box FCN & White-box FCN \\
    \hline
    Black-box 1-NN DTW & 7.029E-01 & \cellcolor[rgb]{ .776,  .937,  .808}\textcolor[rgb]{ 0,  .38,  0}{1.843E-02} & 1.559E-01 \\
    \hline
    White-box 1-NN DTW & \cellcolor[rgb]{ 0,  0,  0} & \cellcolor[rgb]{ .776,  .937,  .808}\textcolor[rgb]{ 0,  .38,  0}{2.748E-03} & 6.887E-02 \\
    \hline
    Black-box FCN & \cellcolor[rgb]{ 0,  0,  0} & \cellcolor[rgb]{ 0,  0,  0} & 5.694E-01 \\
    \hline
    \end{tabular}%
    \end{adjustbox}
  \label{tab:mse_pvalue}%
\end{table}%

Finally, we test how well GATN generalizes onto an unseen dataset, $D_{test}$, such that GATN does not require any additional training. This is beneficial in situations where the time series adversarial samples are generated in constant time of a single forward pass of the GATN model without requiring further training. Such a generalization is uncommon to adversarial methodologies (Fast Gradient Sign Method or Jacobian-based Saliency Map Attack \cite{yuan2019adversarial}) because they require retraining to generate adversarial samples.  Our proposed methodology is robust, successfully generating adversarial samples on data that is unseen to both the GATN and the student models, for the respective targeted time series classification models. Figure \ref{fig:gen} depicts the number of adversarial samples detected, on an unseen dataset, with a white-box and black-box attack on the 1-NN DTW classifiers and FCN classifiers. The white-box attack on the FCN classifier obtains the most adversarial samples per dataset. This is followed by a white-box and black-box attack on the 1-NN DTW, which show similar number of adversarial samples constructed. Finally, we find that the FCN classifier is the least susceptible to black-box attacks.

The unique consequence of this generalization is the application of trained GATN models for attacks that are feasible on real world devices, even for black box attacks. The deployment of a trained GATN with the paired student model affords a near constant-time cost of generating reasonable number of adversarial samples. As the forward pass of  the GATN requires few resources, and the student model is small enough to compute the input gradient ($\tilde{x}$) in reasonable time, these attacks can be constructed without significant computation on small, portable devices. Therefore, the fact that certain classifiers that are trained on certain datasets can be attacked without requiring any additional on-device training is concerning.

\section{Conclusion \& Future Work}
\label{conclusion}
We propose a model distillation technique to mimic the behavior of the various classical time series classification models and an adversarial transformation network to attack various time series datasets. The proposed methodology is applied onto 1-NN DTW and Fully Connected Network (FCN) that are trained on 42 University of California Riverside (UCR) datasets. All 42 datasets were susceptible to attacks. To the best of our knowledge, such an attack on time series classification models has never been done before. The FCN model is prone to more adverarial attacks than 1-NN DTW. We recommend future researchers that develop time series classification models to consider model robustness as an evaluative metric. Finally, we recommend incorporating adversarial data samples into their training data sets in order to further improve resilience to adversarial attacks.

\bibliographystyle{IEEEtran}

\bibliography{biblio}{}
\clearpage

\pagebreak
\appendix[Detailed Results]
\raggedbottom
\begin{table}[H]
  \centering
  \caption{Black-box attack on FCN models}
    \begin{tabular}{|c|c|c|}
    \hline
    Name  & Num. of Adversaries & MSE \\
    \hline
    Car   & 3     & 0.113 \\
    \hline
    ChlorineConcentration & 402   & 0.269 \\
    \hline
    CinCECGTorso & 54    & 0.049 \\
    \hline
    Earthquakes & 5     & 0.122 \\
    \hline
    ECG200 & 4     & 0.138 \\
    \hline
    ECG5000 & 25    & 0.153 \\
    \hline
    ECGFiveDays & 31    & 0.083 \\
    \hline
    FordA & 103   & 0.113 \\
    \hline
    FordB & 761   & 0.119 \\
    \hline
    InsectWingbeatSound & 93    & 0.158 \\
    \hline
    ItalyPowerDemand & 41    & 0.080 \\
    \hline
    Lightning2 & 10    & 0.133 \\
    \hline
    Lightning7 & 5     & 0.132 \\
    \hline
    MoteStrain & 22    & 0.183 \\
    \hline
    NonInvasiveFetalECGThorax1 & 28    & 0.084 \\
    \hline
    NonInvasiveFetalECGThorax2 & 17    & 0.087 \\
    \hline
    Phoneme & 3     & 0.075 \\
    \hline
    Plane & 8     & 0.273 \\
    \hline
    SonyAIBORobotSurface1 & 72    & 0.168 \\
    \hline
    SonyAIBORobotSurface2 & 5     & 0.139 \\
    \hline
    StarLightCurves & 577   & 0.064 \\
    \hline
    Trace & 0     & 0.078 \\
    \hline
    TwoLeadECG & 318   & 0.108 \\
    \hline
    Wafer & 543   & 0.133 \\
    \hline
    AllGestureWiimoteX & 36    & 0.120 \\
    \hline
    AllGestureWiimoteY & 26    & 0.122 \\
    \hline
    AllGestureWiimoteZ & 35    & 0.109 \\
    \hline
    DodgerLoopDay & 3     & 0.095 \\
    \hline
    DodgerLoopGame & 13    & 0.026 \\
    \hline
    DodgerLoopWeekend & 2     & 0.137 \\
    \hline
    EOGHorizontalSignal & 2     & 0.058 \\
    \hline
    EOGVerticalSignal & 24    & 0.064 \\
    \hline
    FreezerRegularTrain & 182   & 0.149 \\
    \hline
    FreezerSmallTrain & 787   & 0.131 \\
    \hline
    Fungi & 8     & 0.205 \\
    \hline
    GesturePebbleZ1 & 10    & 0.129 \\
    \hline
    GesturePebbleZ2 & 10    & 0.130 \\
    \hline
    PickupGestureWiimoteZ & 1     & 0.135 \\
    \hline
    PigAirwayPressure & 1     & 0.040 \\
    \hline
    PigArtPressure & 1     & 0.038 \\
    \hline
    PigCVP & 1     & 0.041 \\
    \hline
    ShakeGestureWiimoteZ & 2     & 0.122 \\
    \hline
    \end{tabular}%
  \label{tab:fcn_bb}%
\end{table}%

\pagebreak
\begin{table}[htbp!]
  \centering
  \caption{White-box attack on FCN models}
    \begin{tabular}{|c|c|c|}
    \hline
    Name  & Number of Adversaries & MSE \\
    \hline
    Car   & 22    & 0.073 \\
    \hline
    ChlorineConcentration & 991   & 0.092 \\
    \hline
    CinCECGTorso & 321   & 0.067 \\
    \hline
    Earthquakes & 78    & 0.118 \\
    \hline
    ECG200 & 25    & 0.254 \\
    \hline
    ECG5000 & 206   & 0.146 \\
    \hline
    ECGFiveDays & 52    & 0.190 \\
    \hline
    FordA & 796   & 0.091 \\
    \hline
    FordB & 868   & 0.113 \\
    \hline
    InsectWingbeatSound & 267   & 0.143 \\
    \hline
    ItalyPowerDemand & 182   & 0.208 \\
    \hline
    Lightning2 & 7     & 0.124 \\
    \hline
    Lightning7 & 20    & 0.123 \\
    \hline
    MoteStrain & 42    & 0.264 \\
    \hline
    NonInvasiveFetalECGThorax1 & 915   & 0.021 \\
    \hline
    NonInvasiveFetalECGThorax2 & 913   & 0.008 \\
    \hline
    Phoneme & 54    & 0.013 \\
    \hline
    Plane & 44    & 0.243 \\
    \hline
    SonyAIBORobotSurface1 & 123   & 0.140 \\
    \hline
    SonyAIBORobotSurface2 & 287   & 0.201 \\
    \hline
    StarLightCurves & 899   & 0.002 \\
    \hline
    Trace & 1     & 0.117 \\
    \hline
    TwoLeadECG & 368   & 0.290 \\
    \hline
    Wafer & 2805  & 0.420 \\
    \hline
    AllGestureWiimoteX & 51    & 0.088 \\
    \hline
    AllGestureWiimoteY & 224   & 0.106 \\
    \hline
    AllGestureWiimoteZ & 195   & 0.183 \\
    \hline
    DodgerLoopDay & 14    & 0.173 \\
    \hline
    DodgerLoopGame & 25    & 0.172 \\
    \hline
    DodgerLoopWeekend & 49    & 0.120 \\
    \hline
    EOGHorizontalSignal & 120   & 0.049 \\
    \hline
    EOGVerticalSignal & 12    & 0.063 \\
    \hline
    FreezerRegularTrain & 727   & 0.029 \\
    \hline
    FreezerSmallTrain & 606   & 0.012 \\
    \hline
    Fungi & 73    & 0.054 \\
    \hline
    GesturePebbleZ1 & 5     & 0.159 \\
    \hline
    GesturePebbleZ2 & 6     & 0.108 \\
    \hline
    PickupGestureWiimoteZ & 10    & 0.107 \\
    \hline
    PigAirwayPressure & 5     & 0.039 \\
    \hline
    PigArtPressure & 87    & 0.037 \\
    \hline
    PigCVP & 42    & 0.039 \\
    \hline
    ShakeGestureWiimoteZ & 13    & 0.125 \\
    \hline
    \end{tabular}%
  \label{tab:fcn_wb}%
\end{table}%

\pagebreak
\begin{table}[htbp]
  \centering
  \caption{Black-box attack on DTW models}
    \begin{tabular}{|c|c|c|}
    \hline
    Name  & Number of Adversaries & MSE \\
    \hline
    Car   & 9     & 0.058 \\
    \hline
    ChlorineConcentration & 576   & 0.160 \\
    \hline
    CinCECGTorso & 57    & 0.048 \\
    \hline
    Earthquakes & 34    & 0.113 \\
    \hline
    ECG200 & 8     & 0.141 \\
    \hline
    ECG5000 & 108   & 0.183 \\
    \hline
    ECGFiveDays & 60    & 0.172 \\
    \hline
    FordA & 249   & 0.117 \\
    \hline
    FordB & 238   & 0.112 \\
    \hline
    InsectWingbeatSound & 202   & 0.181 \\
    \hline
    ItalyPowerDemand & 206   & 0.145 \\
    \hline
    Lightning2 & 3     & 0.095 \\
    \hline
    Lightning7 & 8     & 0.185 \\
    \hline
    MoteStrain & 46    & 0.334 \\
    \hline
    NonInvasiveFetalECGThorax1 & 758   & 0.101 \\
    \hline
    NonInvasiveFetalECGThorax2 & 69    & 0.002 \\
    \hline
    Phoneme & 65    & 0.074 \\
    \hline
    Plane & 18    & 0.201 \\
    \hline
    SonyAIBORobotSurface1 & 16    & 0.137 \\
    \hline
    SonyAIBORobotSurface2 & 42    & 0.151 \\
    \hline
    StarLightCurves & 752   & 0.313 \\
    \hline
    Trace & 0     & 0.136 \\
    \hline
    TwoLeadECG & 178   & 0.194 \\
    \hline
    Wafer & 1843  & 0.204 \\
    \hline
    AllGestureWiimoteX & 34    & 0.123 \\
    \hline
    AllGestureWiimoteY & 38    & 0.128 \\
    \hline
    AllGestureWiimoteZ & 55    & 0.140 \\
    \hline
    DodgerLoopDay & 10    & 0.193 \\
    \hline
    DodgerLoopGame & 4     & 0.164 \\
    \hline
    DodgerLoopWeekend & 3     & 0.141 \\
    \hline
    EOGHorizontalSignal & 36    & 0.064 \\
    \hline
    EOGVerticalSignal & 33    & 0.065 \\
    \hline
    FreezerRegularTrain & 612   & 0.185 \\
    \hline
    FreezerSmallTrain & 139   & 0.150 \\
    \hline
    Fungi & 33    & 0.172 \\
    \hline
    GesturePebbleZ1 & 16    & 0.134 \\
    \hline
    GesturePebbleZ2 & 10    & 0.127 \\
    \hline
    PickupGestureWiimoteZ & 5     & 0.150 \\
    \hline
    PigAirwayPressure & 5     & 0.038 \\
    \hline
    PigArtPressure & 6     & 0.043 \\
    \hline
    PigCVP & 3     & 0.044 \\
    \hline
    ShakeGestureWiimoteZ & 2     & 0.142 \\
    \hline
    \end{tabular}%
  \label{tab:dtw_bb}%
\end{table}%

\pagebreak
\begin{table}[htbp]
  \centering
  \caption{White-box attack on DTW models}
    \begin{tabular}{|c|c|c|}
    \hline
    Name  & Number of Adversaries & MSE \\
    \hline
    Car   & 9     & 0.316 \\
    \hline
    ChlorineConcentration & 477   & 0.156 \\
    \hline
    CinCECGTorso & 64    & 0.051 \\
    \hline
    Earthquakes & 30    & 0.123 \\
    \hline
    ECG200 & 8     & 0.226 \\
    \hline
    ECG5000 & 102   & 0.196 \\
    \hline
    ECGFiveDays & 68    & 0.231 \\
    \hline
    FordA & 247   & 0.131 \\
    \hline
    FordB & 223   & 0.122 \\
    \hline
    InsectWingbeatSound & 203   & 0.167 \\
    \hline
    ItalyPowerDemand & 152   & 0.161 \\
    \hline
    Lightning2 & 3     & 0.099 \\
    \hline
    Lightning7 & 7     & 0.136 \\
    \hline
    MoteStrain & 45    & 0.134 \\
    \hline
    NonInvasiveFetalECGThorax1 & 760   & 0.092 \\
    \hline
    NonInvasiveFetalECGThorax2 & 759   & 0.097 \\
    \hline
    Phoneme & 63    & 0.074 \\
    \hline
    Plane & 19    & 0.207 \\
    \hline
    SonyAIBORobotSurface1 & 22    & 0.132 \\
    \hline
    SonyAIBORobotSurface2 & 50    & 0.184 \\
    \hline
    StarLightCurves & 591   & 0.449 \\
    \hline
    Trace & 0     & 0.111 \\
    \hline
    TwoLeadECG & 219   & 0.201 \\
    \hline
    Wafer & 1295  & 0.182 \\
    \hline
    AllGestureWiimoteX & 34    & 0.126 \\
    \hline
    AllGestureWiimoteY & 30    & 0.121 \\
    \hline
    AllGestureWiimoteZ & 52    & 0.137 \\
    \hline
    DodgerLoopDay & 10    & 0.188 \\
    \hline
    DodgerLoopGame & 6     & 0.157 \\
    \hline
    DodgerLoopWeekend & 4     & 0.143 \\
    \hline
    EOGHorizontalSignal & 35    & 0.068 \\
    \hline
    EOGVerticalSignal & 33    & 0.063 \\
    \hline
    FreezerRegularTrain & 619   & 0.163 \\
    \hline
    FreezerSmallTrain & 136   & 0.131 \\
    \hline
    Fungi & 35    & 0.158 \\
    \hline
    GesturePebbleZ1 & 15    & 0.125 \\
    \hline
    GesturePebbleZ2 & 9     & 0.133 \\
    \hline
    PickupGestureWiimoteZ & 4     & 0.144 \\
    \hline
    PigAirwayPressure & 3     & 0.039 \\
    \hline
    PigArtPressure & 3     & 0.044 \\
    \hline
    PigCVP & 3     & 0.042 \\
    \hline
    ShakeGestureWiimoteZ & 2     & 0.144 \\
    \hline
    \end{tabular}%
  \label{tab:dtw_wb}%
\end{table}%



%




\ifCLASSOPTIONcaptionsoff
  \newpage
\fi

\end{document}